\documentclass{article}
\usepackage{amsmath,graphicx,mlspconf}
\usepackage{xcolor}

\usepackage{amsfonts}
\usepackage{booktabs}

\usepackage{listings}
\usepackage{xcolor}

\copyrightnotice{979-8-3503-2411-2/25/\$31.00 {\copyright}2025 IEEE}

\toappear{2025 IEEE International Workshop on Machine Learning for Signal Processing, Aug.\ 31-- Sep.\ 3, 2025, Istanbul, Turkey}

\title{Tackling Distribution Shift in LLM via KILO: Knowledge-Instructed Learning for Continual Adaptation}
\name{
    Iing Muttakhiroh
    \qquad Thomas Fevens
}

\address{Concordia University, Montreal, Canada}

\begin{document}

\maketitle

\begin{abstract}
Large Language Models (LLMs) often suffer from performance degradation when faced with domain shifts, primarily due to catastrophic forgetting. In this work, we propose KILO (Knowledge-Instructed Learning for Continual Adaptation), a novel continual learning framework that integrates dynamic knowledge graphs with instruction tuning. By leveraging retrieved domain-specific knowledge as guidance during training, KILO enhances both adaptability to new domains and retention of previously acquired knowledge. We pretrain our model on WikiText-103 and evaluate sequential adaptation across four diverse target domains: BioASQ, SciQ, TweetEval, and MIND. Our experiments demonstrate that KILO consistently outperforms strong baselines, including continual fine-tuning, ERNIE 2.0, and CPT, in terms of backward transfer, forward transfer, F1 score, retention rate, and training efficiency. These results highlight the effectiveness of combining structured knowledge retrieval and instruction prompting to overcome domain shift challenges in continual learning scenarios.
\end{abstract}
\begin{keywords}
Large Language Model, Continual Learning, Knowledge Graph, Domain Shift, Instruction Tuning
\end{keywords}

\section{Introduction}
\label{sec:intro}
Large language models (LLMs) have rapidly emerged as a central topic in natural language processing (NLP), demonstrating impressive capabilities across a wide range of tasks. From powering conversational agents to enabling complex question-answering systems, LLMs are increasingly integrated into real-world applications. However, despite their success, a critical challenge remains: \textbf{domain shift} \cite{sarfraz2023error}. When LLMs are exposed to new or specialized domains (e.g., biomedical, legal, or scientific texts), their performance often degrades. This degradation is primarily due to \textbf{catastrophic forgetting}, a phenomenon where the model forgets previously acquired knowledge as it learns new information. As a result, domain shift becomes a major obstacle to achieving robust generalization across diverse and evolving knowledge areas \cite{gupta2023continual,cossu2024continual, qin2023recyclable}.

To address this issue, we explore \textbf{continual learning (CL)} — specifically, the \textbf{replay-based approach}, which is widely regarded as an effective strategy for mitigating catastrophic forgetting \cite{shi2024continual}. To help the model reinforce old knowledge while integrating new information, replay methods function by revisiting and retraining on previously encountered tasks or data. However, this approach's success strongly depends on how well the replay memory is managed, particularly how knowledge is selected, stored, and retrieved efficiently \cite{bagus2021investigation}. One particularly effective solution is using \textbf{knowledge graphs (KGs)}, which have gained widespread attention for their ability to represent domain knowledge in a structured and semantically rich format. A knowledge graph organizes concepts as interconnected entities and relationships, enabling compact storage and meaningful reasoning over the knowledge space. This structured representation enables more informed decision-making during replay—for example, by prioritizing important tasks, identifying knowledge gaps, and ensuring coverage across diverse subdomains \cite{shen2023graph, yuan2023continual}. 

On the other hand, recent studies have shown that \textbf{instruction tuning} significantly improves the generalization capabilities of LLMs by aligning them with task objectives using natural language prompts \cite{gupta2022instructdial, mishra2021cross, mizrahi2024state}. However, instruction tuning alone has limitations: it often lacks grounding in explicit domain knowledge, making it prone to hallucinations or superficial reasoning, especially in specialized or dynamic domains \cite{zhou2024larger, qin2023chatgpt}. Conversely, while knowledge graphs provide explicit, interpretable representations of structured knowledge, they struggle to guide LLMs effectively without additional mechanisms for contextualization and integration into language understanding \cite{zhang2022cglb, zhang2024continual}. 

By encoding both content and context, KGs provide an ideal backbone for managing replay memory in CL setups. When combined with instruction tuning, which guides the model to effectively leverage structured knowledge during inference or fine-tuning, this integration addresses the individual limitations of each approach. Instruction tuning alone may lack grounding in explicit domain knowledge, while knowledge graphs, though rich in semantics, require contextual activation to influence model behaviour. Their synergy enhances both reasoning and generalization, particularly across domain shifts. Motivated by these observations, we propose \textbf{KILO} (Knowledge-Instructed Learning for Continual Adaptation), a novel replay strategy that integrates domain-specific KGs with instruction tuning to improve LLM performance in shifting domains while mitigating catastrophic forgetting. This study is driven by three key research questions: (1) Can explicit knowledge instructions guide adaptation across shifting domains? (2) Does KILO outperform standard fine-tuning and other continual learning baselines under domain shift? Moreover, (3) How does KILO balance adaptability and knowledge retention?

The following sections review related works, formally define our problem setting, describe the proposed KILO framework, and present our experimental results.

\section{Related Works}
\label{sec:related_works}
Several recent works have shaped research in CL, knowledge enhancement, and domain adaptation for language models. Gu {\em et al.}~\cite{gu2022knowledge} propose Knowledge Enhanced Prompt Tuning (KEPT), which retrieves external knowledge to improve few-shot learning, though its reliance on retrieval quality poses risks in noisy domains. Li and Hoiem \cite{li2017learning} introduce Learning without Forgetting (LwF), a foundational continual learning method that preserves old knowledge via soft targets but struggles with significant domain shifts. Sun {\em et al.}~\cite{sun2020ernie} present ERNIE 2.0, a continual pretraining framework using multi-task objectives to mitigate catastrophic forgetting, yet it insufficiently addresses complex, interacting distribution shifts. Jin {\em et al.}~\cite{jin2021lifelong} explore Lifelong Pretraining, adapting models continually to evolving corpora, but which faces stability challenges over long sequences. Similarly, Ke {\em et al.}~\cite{ke2022continual} tackle continual few-shot learning but primarily focus on small data regimes, limiting scalability. Zhou and Cao \cite{zhou2021overcoming} address forgetting in graph neural networks through experience replay, while Zhang {\em et al.}~\cite{zhang2022cglb} introduce the CGLB benchmark for continual graph learning, although both studies emphasize graphs more than textual domains. Pan {\em et al.}~\cite{pan2020unifying} survey unifying knowledge graph learning and reasoning, but stop short of empirical guidance for LLM integration. Yu {\em et al.}~\cite{yu2025graph2text} and Guo {\em et al.}~\cite{guo2024learning} explore graph-to-text learning with LLMs, highlighting the promise of graph-structured knowledge injection, though practical deployments remain immature. Mishra {\em et al.}~\cite{mishra2021cross} advocate using natural language instructions to improve cross-task generalization, suggesting instruction tuning as a key tool for continual adaptation. Gupta {\em et al.}~\cite{gupta2023continual} revisit continual pretraining strategies for LLMs, pointing out that naive continued training often leads to suboptimal performance, needing careful re-warming strategies. Finally, Mizrahi {\em et al.}~\cite{mizrahi2024state} call for richer, multi-prompt evaluation protocols for LLMs, arguing that standard evaluations underestimate models' actual generalization capacities. Despite these advances, challenges remain in maintaining long-term knowledge retention while adapting efficiently to new domains under distribution shifts. To address these limitations, we propose KILO, a framework that integrates dynamic memory management via knowledge graphs and instruction-guided continual optimization, detailed in the following section.

\section{Problem Formulation}
\label{sec:problem_formulation}
\begin{figure}[htb]
    \begin{minipage}[b]{1.0\linewidth}
    \centering
    \centerline{\includegraphics[width=8.5cm]{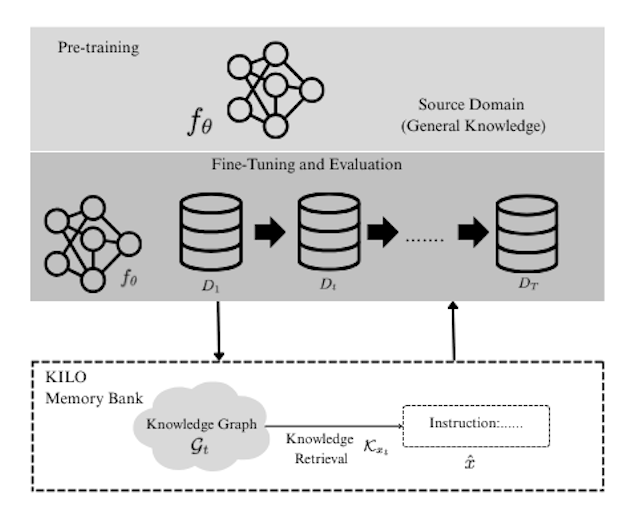}}
    \end{minipage}

    \caption{Overview of the KILO framework. The model is initially pretrained on a general source domain. During fine-tuning and evaluation across sequential target domains $\mathcal{D}_1, \mathcal{D}_2,...\mathcal{D}_T$, KILO maintains a dynamic memory bank implemented as a knowledge graph $\mathcal{G}_t$. At each task, relevant knowledge $\mathcal{K}_{x_t}$ is retrieved based on the current input and reformulated into an instruction $\hat{x}$, guiding continual adaptation while preserving previously acquired knowledge to mitigate forgetting.}
    \label{fig:kilo}
\end{figure}
We address the problem of domain shift in LLMs under a CL setting. Specifically, we consider a sequence of evolving data domains $\mathcal{D}_1,\mathcal{D}_2,...,\mathcal{D}_T$, where each $\mathcal{D}_t$ represents a new domain encountered by the model at time step $t$. These domains may differ in topic, writing style, terminology, or intent distribution. The topics considered are biomedical QA, science-related questions, social media language, and evolving event/topic language. Our goal is to enable a pretrained LLM $f_\theta$ to adapt to each new domain while preserving performance on previously seen domains $(\mathcal{D}_{<t})$ and avoiding catastrophic forgetting, also at the same time, maintaining computational efficiency. To this end, we propose KILO, which leverages a dynamically updated knowledge graph as a memory bank and uses instruction tuning to inject structured prior knowledge into the model via prompting. There are four components in this approach: (1) continual adaptation across domains \cite{jin2021lifelong}, (2) dynamic memory bank via knowledge graph \cite{pan2020unifying}, (3) knowledge-instructed prompt injection, and (4) continual optimization with logit distillation. We illustrate the overall workflow of our proposed KILO framework in Figure \ref{fig:kilo}, detailing the integration of continual fine-tuning with dynamic knowledge injection to overcome domain shift.

The first component is the continual adaptation across domains. Let $x_t \in \mathcal{D}_t$ denote an input at time step $t$; we examine the model's ability to generalize across a sequence of $T$ domains by optimizing
\begin{equation}
    \textup{min}_\theta \sum _{t=1}^T \mathbb{E}_{(x_i,y_i)\sim\mathcal{D}_t}\mathcal{L}(f_{\theta_t}(x_t),y_t),
\end{equation}
subject to minimizing performance degradation on $\mathcal{D}_{<t}$ and efficient training without replaying all past data. To address this, we store domain-relevant knowledge in a structured and compressed memory format, which is used to inform the model about prior experience. 

The second component is the core of KILO—a dynamic memory bank implemented as a knowledge graph, denoted as $\mathcal{G}_t$. This graph stores accumulated knowledge from all previously encountered domains and expands over time, enabling structured retention and semantic generalization. This step involves two main components: designing the structure of the knowledge graph and updating the memory bank accordingly. Each entity $e_i$ and relation $r_{ij}$ is stored as a triple:
\begin{equation}
    \mathcal{G}_t = \{(e_i,r_{ij},e_j)| \textup{ extracted from } \mathcal{D}_{\leq t}\}.
\end{equation}

This knowledge graph is a compact representation of prior knowledge, reducing the need to replay entire past datasets. The next substep focuses on updating the memory bank with new domain information while preserving previously learned knowledge. We use Graph Attention Networks (GATs) for the embedding technique to store and update entity representations efficiently. GATs introduce attention mechanisms over neighbouring nodes, allowing the model to dynamically weight relevant knowledge—an essential capability for adapting to domain shifts. After training on $\mathcal{D}_t$, we extract new knowledge triples using entity and relation extraction and merge and duplicate using entity resolution and embedding similarity, respectively,
\begin{equation}
    \begin{aligned}
        \textup{Extract} (\mathcal{D}_t) = \{(e_i,r_{ij},e_j\} \textup{ and } \\        
        \mathcal{G}_t = \mathcal{G}_{t-1} \cup \textup{Merge(Extract(}\mathcal{D}_t)).\\
    \end{aligned}         
\end{equation}

The next component is knowledge-instructed prompt injection. When the model receives a new input $x_t$, KILO utilizes the knowledge graph (KG) to retrieve prior knowledge and formulate an instruction-style prompt. First, we perform entity linking to identify mentions in $x_t$ that exist in $\mathcal{G}_t$, followed by retrieving the relevant neighbourhood. The retrieved triples are then transformed into natural language instructions. For example, given the triple: ("insulin", "used for", "diabetes"), the resulting prompt would be: "Instruction: Remember that insulin is used to treat diabetes". This structure allows the LLM to condition its prediction on structured memory without retraining old data. The retrieval process and the final input to the model can be described by the following mathematical formulation, respectively,
\begin{equation}
    \begin{aligned}
        \mathcal{K}_{x_t} = \{(e_i,r_{ij},e_j) \in \mathcal{G}_t | e_i \in x_t, \textup{and }d(e_i,e_j)\leq k \} \\       
        \textup{and } \hat{x} = \textup{Prompt} (\mathcal{K}_{x_t})+ x_t .
    \end{aligned}
\end{equation}

The final component is continual optimization with logit distillation to mitigate further forgetting. At each step $t$, we maintain a frozen model $f_{\theta_{t-1}}$ and minimize the divergence between its outputs and those of the updated model, ensuring the model does not deviate significantly on shared concepts across domains:
\begin{equation}
    \mathcal{L}_{distill} = \textup{KL}(f_{\theta_t}(x_t)||f_{\theta_t-1}(x_t)).
\end{equation}

\section{Experimental Result and Discussion}
\label{sec:exp_res}
We design experiments to evaluate KILO under realistic continual domain shift conditions. We use two main types of datasets: one for the source domain in the pretraining phase and the other for the target domains that represent shifting distributions. For the source domain, we pretrain the model on WikiText-103 \cite{merity2016pointer}, a large-scale general knowledge corpus that enables the model to acquire a broad understanding of language and world facts. For the target domains, we sequentially fine-tune the model on four datasets that simulate different domain shifts: BioASQ for biomedical question answering \cite{krithara2023bioasq}, SciQ for science-related questions \cite{welbl2017crowdsourcing}, TweetEval for social media language classification \cite{barbieri2020tweeteval}, and MIND for evolving news event understanding \cite{wu2020mind}. This progression introduces increasingly distinct domain characteristics, allowing us to simulate a realistic setting where the language distribution evolves over time.

We choose T5-base \cite{raffel2020exploring} as the backbone model because of its flexibility in handling various tasks under a text-to-text framework. To benchmark KILO, we compare it against three baselines: (1) continual fine-tuning, where the model is sequentially trained across domains without mechanisms to mitigate forgetting; (2) ERNIE 2.0 \cite{sun2020ernie}, which employs continual pretraining with multi-task objectives; (3) CPT \cite{ke2022continual}, a method designed for efficient continual pretraining; and (4) KILO, our proposed knowledge-instructed learning method.

\begin{table*}[t]
    \centering
    \begin{tabular}{l|cccc|cccc}
    \toprule
    Method & \multicolumn{4}{c}{Forward Transfer ($\rightarrow$)} & \multicolumn{4}{c}{Backward Transfer ($\leftarrow$)}\\
    & $\mathcal{D}_{Bio}$ & $\mathcal{D}_{Sci}$ & $\mathcal{D}_{Tweet}$ & $\mathcal{D}_{News}$ & $\mathcal{D}_{Bio}$ & $\mathcal{D}_{Sci}$ & $\mathcal{D}_{Tweet}$ & $\mathcal{D}_{News}$ \\
    \hline
    Continual Fine-tuning & 68.42 & 70.65 & 76.23 & 74.73 & -6.24 & -5.83 & -4.52 & -5.35 \\
    \hline
    Ernie 2.0\cite{sun2020ernie} & 78.53 & 80.75 & 85.43 & 83.25 & 5.25 & 4.84 & 6.53 & 5.92 \\
    \hline
    CPT\cite{ke2022continual} & 77.24 & 78.92 & 84.23 & 82.34 & 3.82 & 3.54 & 4.73 & 4.25 \\
    \hline
    \textbf{KILO (Ours)} & \textbf{82.75} & \textbf{84.93} & \textbf{89.32} & \textbf{86.15} & \textbf{7.24} & \textbf{6.53} & \textbf{8.45} & \textbf{7.85} \\
    \hline
    \end{tabular}
    \vspace{0.2cm}
    \caption{Performance Comparison (\%) of Backward and Forward Transfer Across Domains.}
    \label{tab:comparison_backwardforward} 
\end{table*} 

\begin{table*}[t]
    \centering
    \begin{tabular}{l|c|c|c|c}
    \toprule
    Method &  F1 (\%)  & KR (\%) & TT (\%) & Total (\%) \\
    \hline
    Continual Fine-tuning  & 72.54 & 59.85 & 68.25 & 66.88 \\
    \hline
    Ernie 2.0\cite{sun2020ernie} & 81.94 & 82.15 & 80.52 & 81.54 \\
    \hline
    CPT\cite{ke2022continual}  & 80.64 & 78.53 & 79.87 & 79.68 \\
    \hline
    \textbf{KILO (Ours)} & 86.54 & 88.73 & 89.25 & 88.17 \\
    \hline
    \end{tabular}
    \vspace{0.2cm}
    \caption{Comparison of Average (\%) Model Performance: F1 Score (F1), Knowledge Retention Rate (KR), and Training Time (TT).}
    \label{tab:comparison_all}
\end{table*}

The training procedure follows a two-phase process. In the first phase, we pretrain the T5-base model \cite{raffel2020exploring} on WikiText-103. In the second phase, we sequentially fine-tune the model on the target domains following the order: BioASQ → SciQ → TweetEval → MIND. After completing training on each domain, we evaluate the model on both the current domain to measure adaptation and all previously seen domains to measure retention. Throughout the process, KILO leverages a dynamic memory bank organized as a knowledge graph that stores extracted salient knowledge from each domain. Before training on a new domain, KILO retrieves relevant knowledge from this memory and injects it via instruction prompts to guide adaptation. This mechanism is designed to enhance knowledge retention while promoting fast adaptation to new distributions.

We report five key metrics to comprehensively assess model performance in evaluating CL under domain shifts. Backward transfer (BWT) captures the impact of learning new tasks on previously learned knowledge; positive BWT values are desirable, indicating that new learning improves or at least preserves prior knowledge, while negative BWT indicates forgetting. Forward transfer (FWT) measures how prior learning facilitates performance on future tasks before training on them, with a higher FWT suggesting stronger generalization capabilities. The F1 score measures the balance between precision and recall across tasks, with a higher F1 score indicating better predictive quality. Knowledge retention rate quantifies the extent to which the model preserves performance on earlier domains after adapting to new domains, where a higher retention rate reflects better resistance to catastrophic forgetting.  Finally, training efficiency is assessed based on cumulative training time, where a lower training time signifies faster and more resource-efficient adaptation. In addition to the principal evaluations, we also conduct an ablation study comparing the model’s performance with and without the integration of knowledge graphs and instruction prompts. This analysis highlights the specific contribution of external knowledge and instruction tuning to adaptation, retention, and overall CL effectiveness.

\begin{table*}[t]
    \centering
    \begin{tabular}{l|cccc|cccc}
    \toprule
    Method & \multicolumn{4}{c}{Forward Transfer ($\rightarrow$)} & \multicolumn{4}{c}{Backward Transfer ($\leftarrow$)}\\
    & $\mathcal{D}_{Bio}$ & $\mathcal{D}_{Sci}$ & $\mathcal{D}_{Tweet}$ & $\mathcal{D}_{News}$ & $\mathcal{D}_{Bio}$ & $\mathcal{D}_{Sci}$ & $\mathcal{D}_{Tweet}$ & $\mathcal{D}_{News}$ \\
    \hline
    \textbf{KILO (Ours)} & \textbf{82.75} & \textbf{84.93} & \textbf{89.32} & \textbf{86.15} & \textbf{7.24} & \textbf{6.53} & \textbf{8.45} & \textbf{7.85} \\
    \hline
    w/o Knowledge Graph (KG) & 75.34 & 76.82 & 82.15 & 80.53 & -1.85 & -0.62 & 1.24 & -0.42 \\
    \hline
    w/o Prompt & 78.52 & 79.93 & 85.72 & 84.25 & 3.15 & 2.83 & 5.24 & 4.52 \\
    \hline
    \end{tabular}
    \vspace{0.2cm}
    \caption{Performance Comparison (\%) of with and without KG and Prompt for Sample Generation.}
    \label{tab:withandwithout}
\end{table*}

Table \ref{tab:comparison_backwardforward} presents the forward and backward transfer results across domains. Continual fine-tuning shows moderate forward transfer, indicating minimal generalization between sequential domains. However, it suffers from strongly negative backward transfer, with retention losses 
up to -6.24\% on $\mathcal{D}_{Bio}$. These results illustrate significant forgetting when the model encounters new domain-specific distributions. ERNIE 2.0 and CPT both substantially improve performance compared to continual fine-tuning, achieving positive backward transfer across all domains, meaning that learning new tasks benefits — rather than harms — previous knowledge to some extent. ERNIE 2.0 generally outperforms CPT by a small margin, especially on scientific and news domains. Most notably, our proposed method, KILO, achieves the highest forward transfer scores across all domains, including 89.32\% on $\mathcal{D}_{Tweet}$ and 86.15\% on $\mathcal{D}_{News}$, indicating superior adaptation to new distributions. Additionally, KILO maintains strong positive backward transfer, with improvements up to 8.45\% on $\mathcal{D}_{Tweet}$, confirming that KILO adapts well to new tasks and enhances prior knowledge representations.

In addition to per-domain transfer, Table \ref{tab:comparison_all} summarizes the overall model performance by averaging F1 score (F1), knowledge retention rate (KR), and training efficiency (TT). Continual fine-tuning achieves the weakest total score of 66.88\%, reflecting its retention and stability struggles across shifts. ERNIE 2.0 and CPT improve significantly over continual fine-tuning, achieving 81.54\% and 79.68\% total scores, respectively, with ERNIE 2.0 leading slightly due to better knowledge retention. However, KILO once again surpasses all baselines, reaching an F1 score of 86.54\%, a retention rate of 88.73\%, and a training efficiency of 89.25\%. These results yield a total performance score of 88.17\%, demonstrating KILO’s balanced strength across adaptability, knowledge preservation, and computational efficiency. Importantly, KILO's advantage is not only in final task performance but also in maintaining faster adaptation and minimizing catastrophic forgetting over sequential learning stages.

To further understand the contribution of KILO’s design components, Table \ref{tab:withandwithout} reports the results of an ablation study. Here, we evaluate two degraded variants: one without KG retrieval and another without instruction Prompts. Removing the KG component significantly reduces both forward and backward transfer, and leads to slight negative backward transfer in some domains (e.g., $\mathcal{D}_{News}$, -0.42\%), suggesting that structured external knowledge plays a crucial role in supporting stable continual adaptation. Removing the Prompt component also weakens performance but causes a minor degradation compared to removing KG alone, highlighting that guided instruction tuning further amplifies the model’s ability to integrate new information effectively. The full KILO model — integrating both KG and Prompt — achieves the strongest results across all metrics, confirming that the synergy between structured knowledge retrieval and instruction guidance is critical to achieving CL success under domain shift conditions. Overall, these experiments confirm that KILO effectively mitigates catastrophic forgetting while promoting proactive knowledge accumulation across evolving domains. By dynamically integrating knowledge graphs and instruction prompts, KILO guides adaptation and preserves prior semantic understanding, highlighting a promising direction for future research in knowledge-augmented CL for LLMs.

\section{Conclusion}
\label{sec:conc}

In this work, we address the critical challenge of domain shift in LLMs by proposing KILO, an innovative CL framework that integrates dynamic knowledge graphs and instruction tuning. Our approach enhances both adaptation to new domains and retention of prior knowledge by leveraging external knowledge structures as instructional prompts during sequential learning. Through extensive experiments across diverse domains—biomedical (BioASQ), scientific (SciQ), social media (TweetEval), and evolving news (MIND)—we demonstrate that KILO consistently outperforms strong baselines, including continual fine-tuning, ERNIE 2.0, and CPT. KILO achieves superior results across key evaluation metrics, including F1 score, backward and forward transfer, retention rate, and training efficiency. Additionally, ablation studies confirm the significant contribution of both knowledge graph integration and instruction prompting to overall performance. These findings highlight the effectiveness of combining structured knowledge retrieval and instruction-driven learning to overcome domain shift challenges in continual adaptation settings. Future work will explore scaling KILO to even larger models and more frequent domain shifts, further enhancing its real-world applicability.


\bibliographystyle{IEEEbib}
\bibliography{kilo}

\clearpage
\appendix
\section*{Appendix}
\section{Training Details}
Here, we present the training details for our continual learning framework, including optimization settings and specifics of the Graph Attention Network (GAT) module used for knowledge graph processing. All models were trained using the AdamW optimizer with a learning rate of $2 \times 10^{-5}$, weight decay of 0.01, and batch size of 8. Each task was trained for 3 epochs, with early stopping based on validation loss. For continual learning, we maintained a memory buffer of 200 samples per task. During replay, 10\% of the mini-batch consisted of replayed exemplars. To process the evolving knowledge graphs, we implemented a 2-layer GAT with 8 attention heads per layer and a hidden size of 128. The input node features were initialized using pre-trained BERT sentence embeddings of the associated entity text spans. Attention scores were computed across all 1-hop neighbours to propagate semantic relevance across the graph. Node updates were performed after each task adaptation, and graphs were stored as DGL heterographs for efficient processing.

\section{Knowledge Graph Details}
This is a step-by-step description of our knowledge graph construction and update pipeline. A pseudocode and process diagram are now included. We build and update task-specific knowledge graphs using the following pipeline:

\noindent\textbf{Entity/Relation Extraction}. First, we use spaCy's NER model and dependency parsing to extract named entities and candidate relations. Relations are filtered using pattern matching and curated verb lists relevant to each task domain. We start by applying Named Entity Recognition (NER) and dependency parsing using spaCy to extract key entities and their semantic relationships from each task's training corpus. Entities are noun phrases or proper names relevant to the domain (e.g., “neural networks”, “WHO”, “climate change”), while relations capture the connections (e.g., “is a part of”, “causes”, “administered by”). We only keep entities with NER confidence scores above a set threshold (e.g., 0.85). Relations are constructed from syntactic dependencies, such as nsubj, dobj, and prep, filtered using a whitelist of allowed relation types for robustness.

\noindent\textbf{Coreference Resolution}. Second, we apply the neuralcoref module to resolve coreferent mentions and link them to consistent entity nodes. Using NeuralCoref, we resolve entity mentions across documents (e.g., “COVID-19” = “the virus” = “it”). This helps avoid duplicate nodes in the graph and ensures all references to the same real-world concept are consolidated.

\noindent\textbf{Graph Construction}. Third, while entities are nodes, relations are labelled directed edges. Nodes are initialized with entity type and context embeddings. The KG is represented as a directed labelled graph $\mathcal{G_t}=(e, r)$, where: $e$ is the nodes (entities) and $r$ is the edges (semantic relations). Each node is initialized with a contextual embedding derived from BERT (using the [CLS] token of its span) and stored in a dictionary with metadata such as entity type, source sentence, and occurrence count. Edges are stored as triples: ($e_i,r_{ij},e_j$), where: $e_i$ is the head entity, $r_{ij}$ is the relation type, and $e_j$ is the tail entity. Only edges with a confidence score above 0.6 are added to reduce noise.

\noindent\textbf{Graph Updates}. Forth, after training on a task, the graph is updated as follows: first step is entity merging: If two nodes’ embeddings have a cosine similarity $> 0.9$, they are merged. The second step is Edge reinforcement: Relations appearing multiple times across documents are weighted more heavily in downstream usage (e.g., for prompt generation). The last step is pruning: unused or low-utility nodes/edges (e.g., degree = 1 and frequency = 1) are periodically pruned to reduce memory usage. Graphs are stored and updated using the Deep Graph Library (DGL), which allows for efficient computation with large-scale, heterogeneous graphs.

\noindent\textbf{Graph Encoding with GAT}. Lastly,for downstream use (e.g., prompt conditioning), the KG is encoded using a 2-layer GAT: each layer uses 8 attention heads, input dimension is 768 (matching BERT output), with a hidden layer of size 128, attention is computed over direct neighbours to learn weighted representations based on semantic importance, node embeddings output by GAT are then used to: select exemplars for replay, enrich instruction prompts with contextual entities/relations and assist domain-aware adaptation.

\textbf{Pseudocode:}
\begin{lstlisting}
def build_graph(documents):
    G = initialize_empty_graph()
    for doc in documents:
        entities = extract_entities(doc)
        coref_map = resolve_coref(doc)
        relations = extract_relations(doc, entities)
        G = update_graph(G, entities, relations, coref_map)
    return G
\end{lstlisting}
\section{Prompt  Quality Evaluation}
In this paper, we introduced a prompt quality evaluation that utilizes both automatic and manual assessments. To assess the effectiveness of instruction prompts, we evaluated their linguistic quality and informativeness using two approaches. The first approach is Automatic Metrics, where we computed BLEU and ROUGE-L scores by comparing generated prompts against expert-written task descriptions in the Natural Instructions dataset. The average BLEU score was 32.6 and ROUGE-L was 51.8, indicating moderate alignment with the ground truth. The second approach is human judgment: a group of 3 annotators rated 50 randomly selected prompts on a scale of 1–5 for clarity, specificity, and task relevance. The average scores were: Clarity (4.2), Specificity (4.0), and Relevance (4.3), suggesting that the prompts are generally well-formed and informative. These evaluations support the claim that our KG-informed prompt generation process produces high-quality instructional cues that enhance model adaptation.

\end{document}